%% file: main.tex
\documentclass[letterpaper, 10 pt, conference]{ieee_format/ieeeconf}  

\IEEEoverridecommandlockouts                              

\usepackage[noadjust]{cite}
\usepackage{threeparttable}
\usepackage{url}
\usepackage{xcolor}
\usepackage{kotex}
\usepackage{graphicx}
\usepackage{amsmath,amssymb,amsfonts}

\usepackage{siunitx}
\usepackage{float}
\usepackage{dsfont}
\usepackage{multicol}
\usepackage{bm}
\usepackage{booktabs}
\usepackage{tabularx}
\let\labelindent\relax
\usepackage{enumitem}
\usepackage{diagbox}
\usepackage{multirow}

\usepackage{cuted}
\usepackage{capt-of}
\usepackage[caption=false, font=footnotesize]{subfig}
\usepackage{algorithm}
\usepackage{algpseudocode}
\usepackage{textcomp}
\usepackage{array}
\usepackage{tikz}
\usetikzlibrary{patterns}
\usepackage{circledsteps}
\usepackage{tcolorbox}
\usepackage{colortbl}
\usepackage{xspace}
\tcbuselibrary{breakable, skins}
\usepackage{changepage}

\newcommand{\BfPara}[1]{{\noindent\bf#1.}\xspace}
\definecolor{myblue}{RGB}{38, 54, 146}
\newcommand*\circled[1]{\Circled[inner color=white, fill color=myblue, outer color=myblue]{\scriptsize{#1}}}

\definecolor{color0}{HTML}{2E86AB}
\definecolor{color1}{HTML}{A23B72}
\definecolor{color2}{HTML}{F18F01}
\definecolor{color3}{HTML}{C73E1D}
\definecolor{color4}{HTML}{6A994E}
\definecolor{color5}{HTML}{7209B7}

\definecolor{bestbg}{RGB}{165, 205, 255}
\definecolor{secondbg}{RGB}{195, 220, 255}
\definecolor{thirdbg}{RGB}{225, 238, 255}

\newcommand{\resbar}[3][white]{
  \cellcolor{#1}\makebox[2.2em][r]{#2\,\%}\hspace{3pt}
  \tikz[baseline={([yshift=-0.6ex]current bounding box.center)}]{
    \path (0,0) rectangle (0.9, 0.22);
    \fill[fill=#3] (0,0) rectangle (#2*0.009, 0.22);
    \fill[pattern=north west lines, pattern color=black!30!#3] (0,0) rectangle (#2*0.009, 0.22);
  }%
}

\def\BibTeX{{\rm B\kern-.05em{\sc i\kern-.025em b}\kern-.08em
    T\kern-.1667em\lower.7ex\hbox{E}\kern-.125emX}}
  
\title{\LARGE \bf
    Distilling Collaborative Dynamics into Latent Space\\for Implicit Coordination in Decentralized Multi-Agent Manipulation
}

\author{Chanyoung Park, Minsung Yoon, Andrew Jeong, and $\text{Sung-eui Yoon}^\dagger$
\thanks{The authors are with the School of Computing at the Korea Advanced Institute of Science and Technology (KAIST), Daejeon, 34141, Republic of Korea. E-mails: \texttt{$\{$chanyoung, minsung.yoon, andrew7447$\}$@kaist.ac.kr, sungeui@kaist.edu.} $^\dagger$Corresponding author.}
}

\begin{document}
    \maketitle
    \thispagestyle{empty}
    \pagestyle{empty}
    
    \begin{strip}
    \vspace{-65pt}
    \centering 
    \includegraphics[width=\linewidth]{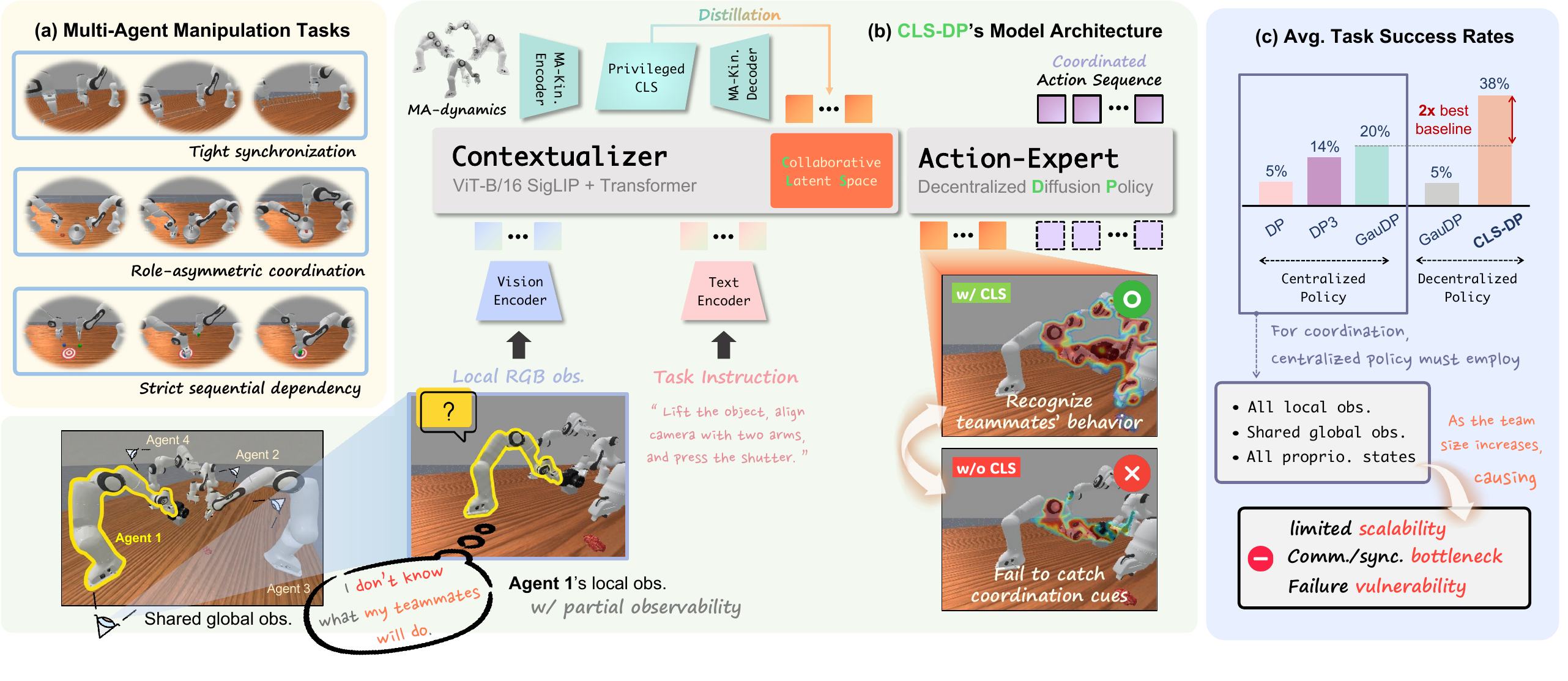}
	\vspace{-25pt}
    \captionof{figure}{\hspace{-8.5pt}
    \textbf{Overview of CLS-DP.}
    (a) Multi-agent manipulation tasks require tight synchronization, role-asymmetric coordination, and strict sequential dependency.
    (b) CLS-DP learns a collaborative latent from privileged multi-agent dynamics in a contextualizer during training. Each agent then infers this latent from its local RGB observation to condition a decentralized diffusion policy at deployment under partial observability.
    Integrated Gradients~\cite{sundararajan2017axiomatic} attribution maps indicate that CLS-DP attends to all agents' joints and grippers, successfully completing the task (top). In contrast, a baseline diffusion policy remains egocentric and fails to complete the task (bottom). This suggests that the collaborative latent helps agents recognize and leverage teammates’ behaviors as implicit coordination cues.
    (c) CLS-DP achieves a 38\% average success rate in various types of multi-agent manipulation tasks, nearly doubling the performance of the best centralized baseline (20\%) without reliance on shared global views, explicit state information, or inter-agent communication.
    }
    \vspace{-9pt}
	\label{fig:overview}
    \end{strip}

    \input{Paper/0_abs}
    \input{Paper/1_intro}
    \input{Paper/2_related_work}
    \input{Paper/3_preliminaries}
    \input{Paper/4_method}
    \input{Paper/5_experiment}
    \input{Paper/6_conclusion}
    
    \section*{ACKNOWLEDGMENT}
    \footnotesize
        This work was supported by the Institute of Information \& Communications Technology Planning \& Evaluation (IITP) grant funded by the Korea government (MSIT) (RS-2023-00237965, Recognition, Action and Interaction Algorithms for Open-world Robot Service), the IITP-ITRC (Information Technology Research Center) grant funded by the Korea government (Ministry of Science and ICT) (IITP-2026-RS-2020-II201460), and the InnoCORE program of the Ministry of Science and ICT (N10260099).
    {
        \small
        \bibliographystyle{ieee_format/ieee}
        \bibliography{./ref}
    }
    
\end{document}

%% file: Paper/0_abs.tex
\begin{abstract}
Multi-arm manipulation demands precise spatiotemporal coordination, yet many centralized approaches scale poorly as team size increases.
To address this, we propose CLS-DP, a decentralized multi-agent framework that enables implicit coordination under partial observability without shared global views, explicit state information, or inter-agent communication.
Under the centralized training and decentralized execution (CTDE) paradigm, CLS-DP distills privileged multi-agent dynamics into a latent space.
At deployment, each agent infers a collaborative latent from its local RGB observation and a shared task instruction; it then conditions the diffusion denoising process on this latent.
This design enables implicit coordination with a per-agent cost independent of team size.
Across six RoboFactory benchmark tasks spanning two to four agents, CLS-DP achieves a 38\% mean success rate, outperforming the best centralized baseline (20\%) and a decentralized ablation without the collaborative latent (9\%). It also maintains superior parameter efficiency across all agent configurations.
Attribution maps show that an agent conditioned on the collaborative latent places high attribution on the joints and grippers of both itself and its teammates throughout execution. This suggests that the learned latent efficiently encodes collaborative dynamics from local observation, which facilitates implicit coordination in realistic settings characterized by partial observability.
\end{abstract}

%% file: Paper/1_intro.tex
\section{Introduction}
\label{sec:1}
Across industrial, medical, and domestic service domains, robotic manipulation is progressively evolving from isolated single-arm operation toward tightly coordinated collaboration among multiple arms~\cite{worldrobotics2025, tang2024automate, kim2024srt, nasiriany2024robocasa, chi2024umi}.
In such settings, robots need to cooperate under spatial, temporal, and physical constraints, where minor misalignment often leads to task failure.
Recent advances in generative policy learning---particularly diffusion-based policies~\cite{chi2024diffusion, song2019generative, wang2022diffusion}---have demonstrated strong capability in modeling high-dimensional, multimodal action distributions for complex manipulation. 
However, most existing multi-arm approaches rely on centralized execution, in which a unified policy coordinates all agents by aggregating their observations and actions.
As the team size increases, this formulation suffers from rapid growth of the joint observation–action space, heightened synchronization complexity, vulnerability to single-agent failures, and dependence on communication and shared observations that are difficult to guarantee in real-world deployments (refer to Sec.~\ref{sec:2-A}).

To handle these limitations, recent work has shifted toward decentralized execution, casting multi-arm manipulation as a coordination problem among multiple agents rather than relying on a unified policy. 
Under this paradigm, each robot executes an independent policy conditioned solely on local observations. 
However, decentralization introduces distinct challenges: the need for consistent coordination under partial observability, where asymmetric and incomplete observations obscure global state information and allow minor temporal or spatial discrepancies to disrupt task progress, often leading to collisions, object drops, or deadlocks.
To alleviate this, many existing approaches rely on assumptions---such as full state access or shared global views---which limit applicability in realistic settings characterized by partial observability and the absence of communication (refer to Sec.~\ref{sec:2-B}).

To this end, we propose CLS-DP (\textbf{C}ollaborative \textbf{L}atent \textbf{S}pace–conditioned \textbf{D}iffusion \textbf{P}olicy), a framework that enables implicit coordination among multiple agents under decentralized execution and partial observability settings without shared global views, explicit state information, or inter-agent communication (refer to Fig.~\ref{fig:overview}).
Under the centralized training and decentralized execution (CTDE) paradigm, CLS-DP learns a latent space that captures collaborative dynamics by distilling privileged multi-agent trajectories available during training.
To enable each agent to leverage this representation under partial observability at deployment, we introduce an agent-specific contextualizer that infers the collaborative intent from its local observation and a shared task instruction.
This design enables agents to recognize teammates’ behaviors from local information alone, thereby improving coordination under realistic partial observability.

We evaluate CLS-DP on the RoboFactory multi-arm manipulation tasks~\cite{qin2025robofactory} with two to four agents requiring structured and role-specific coordination. 
CLS-DP achieves higher success rates than centralized and decentralized baselines, while remaining parameter-efficient as team size grows.

%% file: Paper/2_related_work.tex
\section{Related Work}
\label{sec:2}
\subsection{Policy Learning for Robotic Manipulation}
\label{sec:2-A}
Direct policy learning on hardware is costly and safety-critical, motivating behavior cloning (BC) from demonstrations, which learns a direct mapping from observations to actions~\cite{dalal2023imitating,yin2024offline,ma2024contrastive,foster2024is}.
Early BC approaches represented policies using deterministic regression or explicit parametric distributions, such as Gaussian mixture models; however, their limited expressiveness often leads to mode averaging when modeling the highly multimodal action distributions common in robotic manipulation.
More recent works address this limitation by employing expressive generative models that better capture complex multimodal action distributions.
Energy-based models~\cite{florence2021implicit,du2021improved,ta2022conditional} represent policies as unnormalized energy functions and generate actions via iterative optimization; however, they require contrastive training with negative sampling, which often leads to unstable training dynamics.
Diffusion-based policies~\cite{chi2024diffusion,song2019generative,wang2022diffusion} instead learn the conditional action score function through a denoising objective, enabling more stable training and expressive multimodal action generation without explicit normalization.

While these approaches are mainly studied in single-agent settings with centralized execution~\cite{dong2025mimicd}, many tasks require multi-robot coordination. 
As the number of robots increases, centralized formulations scale poorly due to the growth of the joint observation–action space and synchronization issues~\cite{dong2025mimicd, amato2024initial}.
We instead adopt a decentralized formulation and enable each agent to infer teammates' intent from its local observation, facilitating effective collaboration.

\subsection{Decentralized Multi-Agent Coordination}
\label{sec:2-B}
Recent work has explored decentralized diffusion policies for multi-agent coordination~\cite{zhu2024madiff, dong2025mimicd, he2025latent}. 
MIMIC-D~\cite{dong2025mimicd} adopts a CTDE framework to address the scalability limitations of fully centralized policies and the coordination failures in fully decentralized training. 
Under this paradigm, coordination emerges implicitly through joint supervision over multi-agent demonstrations, without explicitly modeling other agents’ internal states or intentions.
In contrast, MADIFF~\cite{zhu2024madiff} also follows CTDE but forces each agent to predict other agents’ future states by incorporating cross-agent attention mechanisms. 
While such explicit modeling enhances inter-agent awareness, its complexity grows with team size, thereby inheriting scalability issues similar to centralized policies. 
Moreover, both approaches assume near-full state observability, including access to other agents’ positions, whereas RGB-based manipulation must jointly address state inference, coordination, and uncertainty arising from occlusion and limited sensor field of view, all of which increase problem complexity.

LatentToM~\cite{he2025latent} proposes a shared embedding space for consensus alignment among agents; however, since this space is constructed using images captured by a third-person global-view camera, agents rely on the same global observations during both training and inference. 
Moreover, this representation captures current scene state rather than prospective collaborative dynamics among agents.

To address these limitations, we adopt CTDE to distill prospective multi-agent dynamics from future joint demonstration trajectories into an observation-conditioned latent space via a residual conditional variational autoencoder (CVAE).
At decentralized deployment, each agent samples the latent from the learned collaborative latent space and conditions its diffusion policy on it to infer teammates' intents from local RGB observation without reliance on shared global views, explicit state information, or inter-agent communication, while keeping a per-agent inference cost constant as team size grows.

\begin{figure*}[t!]
    \centering
    \captionsetup[subfigure]{labelformat=empty}
    \centering
    \includegraphics[width=\linewidth]{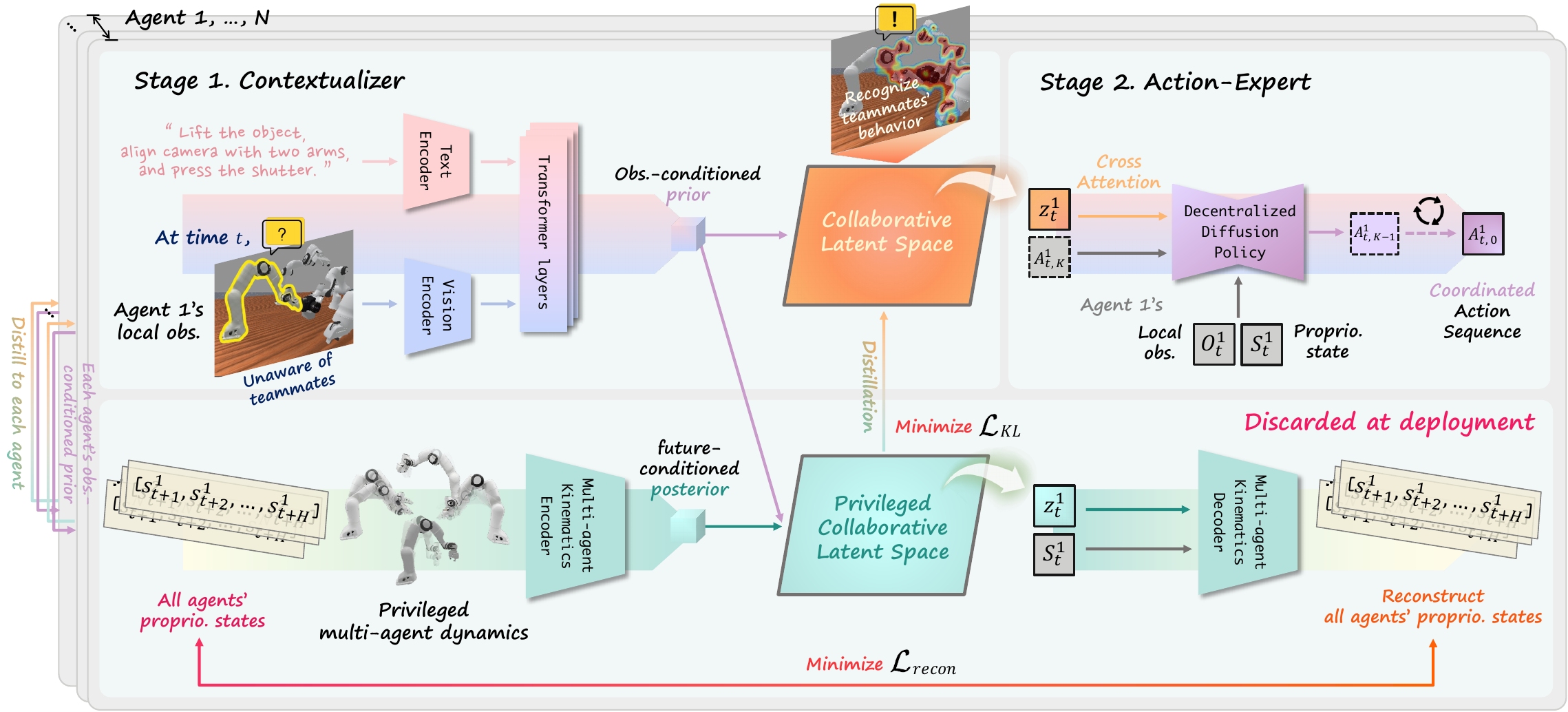}
    \caption{\hspace{-5pt}\textbf{CLS-DP} architecture with two training stages: \textbf{Stage 1. Contextualizer:} at each timestep $t$, a cross-modal prior network encodes the agent $i$'s local RGB observation and shared task instruction into an observation-conditioned prior, while a multi-agent kinematics encoder infers a future-conditioned posterior from privileged joint dynamics of all agents; KL regularization aligns the posterior to the prior so that the kinematics encoder and decoder are discarded at deployment. \textbf{Stage 2. Action-Expert:} the latent $\mathbf{z}^i_t$ sampled from the collaborative latent space is injected via cross-attention into an agent $i$'s decentralized diffusion policy alongside its own local observation and proprioceptive history to generate a coordinated action sequence.}
    \label{fig:sys_pipeline}
    \vspace{-15pt}
\end{figure*}

\subsection{Representation Learning for Robotic Manipulation}
Learning latent representations is crucial for scaling manipulation under high-dimensional, partially observable inputs.
Latent dynamics models encode state transitions in compressed spaces, enabling imagination-based planning and improved sample efficiency~\cite{hansen2024tdmpc2,zhou2025dinowm,cui2024dynamo,hansen2025hierarchical}.
In multi-agent settings, agents co-adapt through interaction, requiring representations that capture inter-agent dependencies and prospective coordination rather than passive scene dynamics. 
We address this challenge by learning a collaborative latent space via distillation of privileged multi-agent state trajectories into an observation-conditioned prior during centralized training. 
At deployment, each agent infers this latent from local observation and conditions its diffusion policy on it, enabling decentralized anticipation of teammates’ behaviors without explicit state sharing or communication.

%% file: Paper/3_preliminaries.tex
\section{Preliminaries}
\label{sec:3}
\BfPara{Diffusion Model}
Denoising diffusion probabilistic models (DDPMs)~\cite{ho2020denoising} learn a distribution by corrupting samples with Gaussian noise over $K$ steps and training a neural network to reverse it. The forward process corrupts a clean sample $\mathbf{x}_0$ into a noisy sample $\mathbf{x}_k$ through a Markov chain:
\begin{equation}
  \mathbf{x}_k = \sqrt{\bar{\alpha}_k}\,\mathbf{x}_0
  + \sqrt{1-\bar{\alpha}_k}\,\boldsymbol{\epsilon}, \quad
  \boldsymbol{\epsilon}\sim\mathcal{N}(\mathbf{0},\mathbf{I}),
  \label{eq:forward}
\end{equation}
where $\bar{\alpha}_k=\prod\nolimits_{i=1}^k\alpha_i$ and $\alpha_k\in(0,1)$ defines a noise schedule that decreases monotonically over $K$ steps.
A network $\boldsymbol{\epsilon}_\Phi$ is trained to predict the injected noise by minimizing:
\begin{equation}
  \mathcal{L}_{\text{Diffusion}}
  = \mathbb{E}_{\mathbf{x}_0,\boldsymbol{\epsilon},k}\!\left[
    \left\|\boldsymbol{\epsilon}
    - \boldsymbol{\epsilon}_\Phi\!\left(\mathbf{x}_k, k\right)
    \right\|^2_2\right].
  \label{eq:diff_loss}
\end{equation}
At inference, a clean sample is recovered by iteratively denoising from $\mathbf{x}_K\sim\mathcal{N}(\mathbf{0},\mathbf{I})$ via the reverse process:
\begin{equation}
    \mathbf{x}_{k-1} = \frac{1}{\sqrt{\alpha_k}}\left(\mathbf{x}_k - \frac{1-\alpha_k}{\sqrt{1-\bar{\alpha}_k}}\boldsymbol{\epsilon}_\Phi(\mathbf{x}_k, k)\right) + \eta_k\boldsymbol{\epsilon}',
    \label{eq:reverse}
\end{equation}
where $\eta_k$ is the reverse process noise scale and $\boldsymbol{\epsilon}'\sim\mathcal{N}(\mathbf{0},\mathbf{I})$.

\BfPara{Diffusion Policy}
Diffusion policy~\cite{chi2024diffusion} instantiates DDPM for expressive and robust visuomotor control by treating the robot action sequence as the generative target.
Given a length-$L$ observation history $\mathbf{O}_t := \mathbf{o}_{t-L+1:t} \in \mathbb{R}^{L\times d_{\mathbf{o}}}$ and proprioceptive history $\mathbf{S}_t := \mathbf{s}_{t-L+1:t} \in \mathbb{R}^{L\times d_{\mathbf{s}}}$, the policy learns the conditional distribution over an action sequence $\mathbf{A}_t := \mathbf{a}_{t:t+H-1} \in \mathbb{R}^{H\times d_{\mathbf{a}}}$, where $H$ is the prediction horizon.\footnote{Hereafter, $d_{*}$ is the dimensionality of the corresponding vector space.}
A denoising network $\boldsymbol{\epsilon}_{\Phi}$ is trained by minimizing:
\begin{equation}
  \mathcal{L}_{\text{DP}}
  = \mathbb{E}_{\mathbf{A}_t,\boldsymbol{\epsilon},k}\!\left[
    \left\|\boldsymbol{\epsilon}
    - \boldsymbol{\epsilon}_{\Phi}\!\left(\mathbf{A}_{t,k}, k, \mathbf{O}_t, \mathbf{S}_t\right)
    \right\|^2_2\right].
  \label{eq:dp_loss}
\end{equation}
Here, $\mathbf{A}_{t,k}$ is the noisy action sequence at diffusion step $k$.
In execution, the clean action sequence $\mathbf{A}_{t,0}$ is sampled via the $K$-step reverse diffusion process, yielding smooth action trajectories sampled from expressive multimodal distributions.

%% file: Paper/4_method.tex
\section{Method}
\label{sec:4}

\subsection{Problem Formulation}
We formulate multi-agent manipulation as a decentralized partially observable Markov decision process (Dec-POMDP)~\cite{oliehoek2016concise}, defined by $\langle N,\mathcal{S},\mathcal{O},\mathcal{A},P,Z,\mathcal{Q}_0\rangle$.
In our decentralized setting, $N$ agents indexed by $i \in \{1,\dots,N\}$ execute their policies conditioned on their own observation and proprioceptive history, without access to shared global RGB observations, explicit global state information, or inter-agent communication at inference time.
Here, $\mathcal{S}=\prod_{i=1}^N\mathcal{S}^i$, $\mathcal{O}=\prod_{i=1}^N\mathcal{O}^i$, and $\mathcal{A}=\prod_{i=1}^N\mathcal{A}^i$ denote the joint state, observation, and action spaces, where $\mathcal{S}^i\subseteq\mathbb{R}^{d_{\mathbf{s}}}$, $\mathcal{O}^i\subseteq\mathbb{R}^{d_{\mathbf{o}}}$ and $\mathcal{A}^i\subseteq\mathbb{R}^{d_{\mathbf{a}}}$.
The transition function is $P:\mathcal{S}\times\mathcal{A}\rightarrow\mathcal{P}(\mathcal{S})$ and the observation function for agent $i$ is
$Z^i:\mathcal{S}\rightarrow\mathcal{P}(\mathcal{O}^i)$, where $\mathcal{P}(\cdot)$ denotes the set of probability measures.
The initial state $\mathbf{s}_0 \sim \mathcal{Q}_0$ sets the environment and robot configurations; $\mathcal{Q}_0$ introduces variability across episodes through randomization of initial configurations.

When the environment is in state $\mathbf{s}_t \in \mathcal{S}$ at time $t$, given the shared task instruction $l$, each agent $i$ receives a local image observation $\mathbf{o}^i_t \in \mathcal{O}^i$ along with its proprioceptive state $\mathbf{s}^i_t \in \mathcal{S}^i$. Each agent then generates an action sequence $\mathbf{A}^i_t \in (\mathcal{A}^i)^H$ according to a decentralized policy $\pi^i(\mathbf{A}^i_t|\mathbf{O}^i_t,\mathbf{S}^i_t,l)$, which is executed for $H$ steps.
At each timestep $u\in\{t, \ldots, t+H-1\}$, the joint action $\mathbf{a}^{1:N}_u\in\mathcal{A}$ induces the environment state transition $\mathbf{s}_{u+1}\sim P(\mathbf{s}_u,\mathbf{a}^{1:N}_u)$.

\subsection{Overview}
\label{sec:4-B}
As in Fig.~\ref{fig:sys_pipeline}, CLS-DP follows a dual-module architecture under the centralized training and decentralized execution (CTDE) paradigm. It consists of a \textbf{contextualizer} that infers a coordination-aware latent variable from its local observation and the task instruction, and an \textbf{action-expert} that generates an action sequence conditioned on this latent.

\subsubsection{Centralized Training}
We first train the contextualizer with privileged multi-agent information to distill collaborative dynamics into an observation-conditioned latent space.
We then freeze it and train a diffusion-based action-expert for each agent conditioned on the inferred latent. Concretely, we optimize the following losses in two stages:
\vspace{-3pt}
\begin{align}
\text{Stage 1:} \quad & \sum_{i=1}^{N} \mathcal{L}_{\text{CT}}(\theta^i, \psi^i), \\
\text{Stage 2:} \quad & \sum_{i=1}^{N} \mathcal{L}_{\text{AE}}(\phi^i; \text{sg}(\theta^i)),
\end{align}
where $(\theta^i, \psi^i)$ and $\phi^i$ parameterize the contextualizer and action-expert for agent $i$, respectively. In Stage~2, $\theta^i$ is fixed by the stop gradient operator $\text{sg}(\cdot)$, and $\psi^i$ is involved only in Stage~1 for privileged learning.

\BfPara{Contextualizer}
Motivated by CVAEs in motion generation, where a latent variable captures plausible pose transitions and the underlying dynamics~\cite{ling2020character,xue2025leverb}, we introduce a coordination-aware latent variable $\mathbf{z}^i_t$ to summarize prospective multi-agent dynamics.
Specifically, the contextualizer enables each agent to infer teammates' intent from its local observation by employing a CVAE to distill privileged multi-agent dynamics into an observation-conditioned prior.

Each agent's prior network encodes the current local image observation $\mathbf{o}^i_t$ and the shared instruction $l$ with frozen pre-trained SigLIP encoders~\cite{zhai2023sigmoid}. It then fuses the embeddings with Transformer layers~\cite{vaswani2017attention} and outputs $(\mu^i_\rho, \sigma^i_\rho)$ to define the observation-conditioned prior:
\begin{equation}
    p_{\theta^i}(\mathbf{z}^i_t | \mathbf{o}^i_t, l) = \mathcal{N}(\mu^i_\rho, \mathrm{diag}((\sigma^i_\rho)^2)).
\end{equation}
Here, only the current local observation serves as input; conditioning on longer histories can make the prior overly informative, preventing $\mathbf{z}^i_t$ from sufficiently learning collaborative dynamics distilled from the posterior~\cite{alemi2018fixing,xue2025leverb}.

For the posterior network, the multi-agent kinematics encoder $E_{\psi^i}$ and decoder $D_{\psi^i}$ are implemented with Transformer layers to capture cross-agent interactions in privileged joint trajectories.
\textit{i)} The multi-agent kinematics encoder predicts residual parameters $(\mu^i_E, \sigma^i_E)$ from the privileged future joint trajectories $\mathbf{s}^{1:N}_{t+1:t+H}$, forming the residual posterior:
\begin{equation}
    q_{\psi^i}(\mathbf{z}^i_t | \mathbf{s}^{1:N}_{t+1:t+H}) = \mathcal{N}(\mu^i_\rho + \mu^i_E,\, \mathrm{diag}((\sigma^i_E)^2)).
\end{equation}
\textit{ii)} The latent is sampled via the reparameterization trick, $\mathbf{z}^i_t = \mu^i + \sigma^i \odot \boldsymbol{\nu}$, $\boldsymbol{\nu} \sim \mathcal{N}(\mathbf{0}, \mathbf{I})$, where $\mu^i = \mu^i_\rho + \mu^i_E$ and $\sigma^i = \sigma^i_E$.
\textit{iii)} The multi-agent kinematics decoder reconstructs the future joint trajectories from only $(\mathbf{s}^i_t, \mathbf{z}^i_t)$, verifying that $\mathbf{z}^i_t$ encodes sufficient information about prospective multi-agent dynamics beyond the agent's own state.

Together, the contextualizer is trained by minimizing:
\begin{multline}
\mathcal{L}_{\text{CT}}(\theta^i, \psi^i) = \beta\,\underbrace{D_{\mathrm{KL}}\!\left( q_{\psi^i}(\mathbf{z}^i_t | \mathbf{s}^{1:N}_{t+1:t+H}) \,\|\, p_{\theta^i}(\mathbf{z}^i_t | \mathbf{o}^i_t, l) \right)}_{\text{Distill privileged collaborative dynamics into the prior}} \\
+ \underbrace{\mathbb{E}_{\mathbf{z}^i_t \sim q_{\psi^i}\!\left(\mathbf{z}^i_t \mid \mathbf{s}^{1:N}_{t+1:t+H}\right)}\!\left[ \| \hat{\mathbf{s}}^{1:N}_{t+1:t+H} - \mathbf{s}^{1:N}_{t+1:t+H}\|_2^2 \right]}_{\text{Reconstruct privileged future joint trajectories}},
\end{multline}
where $\hat{\mathbf{s}}^{1:N}_{t+1:t+H} \!\!:=\!\! D_{\psi^i}(\mathbf{s}^i_t, \mathbf{z}^i_t)$.
The KL regularizes the residual posterior (centered at $\mu^i_\rho+\mu^i_E$) toward the prior by reducing residual shifts ($\mu^i_E\!\to\!0$) and aligning uncertainty ($\sigma^i_E\!\to\!\sigma^i_\rho$), so that the prior alone suffices at deployment.

\BfPara{Action-Expert}
After training the contextualizer, we train an action-expert conditioned on the collaborative latent $\mathbf{z}^i_t$ sampled from the prior $p_{\theta^i}(\mathbf{z}^i_t | \mathbf{o}^i_t, l)$, without backpropagating gradients into the contextualizer.
Here, $\mathbf{z}^i_t$ is injected into the downsampling and bottleneck stages through cross-attention blocks, where U-Net features preserve multi-scale spatial information and yield stronger control representations than upsampling-layer features~\cite{parisi2022unsurprising,gupta2024pretrained}.
In parallel, $(\mathbf{O}^i_t,\mathbf{S}^i_t)$ are incorporated via FiLM layers~\cite{perez2018film}.

The conditional reverse denoising step is iterated $K$ times:
\begin{multline}
\mathbf{A}^i_{t,k-1}
\!=\!
\lambda_k\Big(
\mathbf{A}^i_{t,k}
-
\gamma_k\,\boldsymbol{\epsilon}_{\phi^i}(\mathbf{A}^i_{t,k}, k, \mathbf{O}^i_t, \mathbf{S}^i_t, \mathbf{z}^i_t)
+
\eta_k \boldsymbol{\epsilon}'
\Big),
\end{multline}
where $\lambda_k$ and $\gamma_k$ are scalar coefficients following the denoising schedule, as with $\eta_k$.
Each agent $i$ trains its denoiser $\boldsymbol{\epsilon}_{\phi^i}$ to predict the injected noise at each diffusion step with:
\begin{multline}
\mathcal{L}_{\text{AE}}
\!=\!
\mathbb{E}_{\mathbf{A}^i_t,\boldsymbol{\epsilon},k} \!
\left[
\left\|
\boldsymbol{\epsilon}
\!-\!
\boldsymbol{\epsilon}_{\phi^i}\!\left(\mathbf{A}^{i}_{t,k}, k,\mathbf{O}^i_t, \mathbf{S}^i_t, \mathrm{sg}(\mathbf{z}^i_t)\right)
\right\|_2^2
\right],\!\!
\end{multline}
where $k$ is sampled uniformly from $\{1,\dots,K\}$.
Conditioned on $\mathbf{z}^i_t$, the action-expert denoises actions from local observation history while capturing collaborative dynamics, thereby generating implicitly coordinated actions.

\subsubsection{Decentralized Execution}
Existing multi-agent diffusion policies typically condition on all agents' histories of observation and proprioceptive state at runtime under a centralized paradigm~\cite{jiang2023motiondiffuser,niedoba2023diffusion,wang2025reinventing}.
In contrast, each agent executes CLS-DP as a two-stage pipeline conditioned on its local observation and proprioceptive history, imposing a per-agent computational cost that remains constant as the team size grows.
First, each agent $i$ samples a collaborative latent variable $\mathbf{z}^i_t \sim p_{\theta^i}(\mathbf{z}^i_t | \mathbf{o}^i_t, l)$ from the contextualizer.
Second, the action-expert performs the $K$-step reverse denoising process conditioned on $(\mathbf{O}^i_t,\mathbf{S}^i_t,\mathbf{z}^i_t)$ to generate an action sequence $\mathbf{A}^i_t$.
In summary, CLS-DP instantiates the decentralized policy at inference time as the marginal distribution:
\begin{multline}
\pi^i_{\theta^i,\phi^i}(\mathbf{A}^i_t | \mathbf{O}^i_t, \mathbf{S}^i_t, l)\\
= \int \tau_{\phi^i}(\mathbf{A}^i_t | \mathbf{z}^i_t, \mathbf{O}^i_t, \mathbf{S}^i_t)
\, p_{\theta^i}(\mathbf{z}^i_t | \mathbf{o}^i_t, l) \, d\mathbf{z}^i_t \,,
\end{multline}
where $\tau_{\phi^i}$ is the action distribution induced by the $K$-step reverse denoising process of agent $i$'s action-expert.

%% file: Paper/5_experiment.tex
\begin{figure}
    \centering
    \subfloat[Failure \circled{1}\,: Object drop from unsynchronized lifting.]{
        \includegraphics[width=\linewidth]{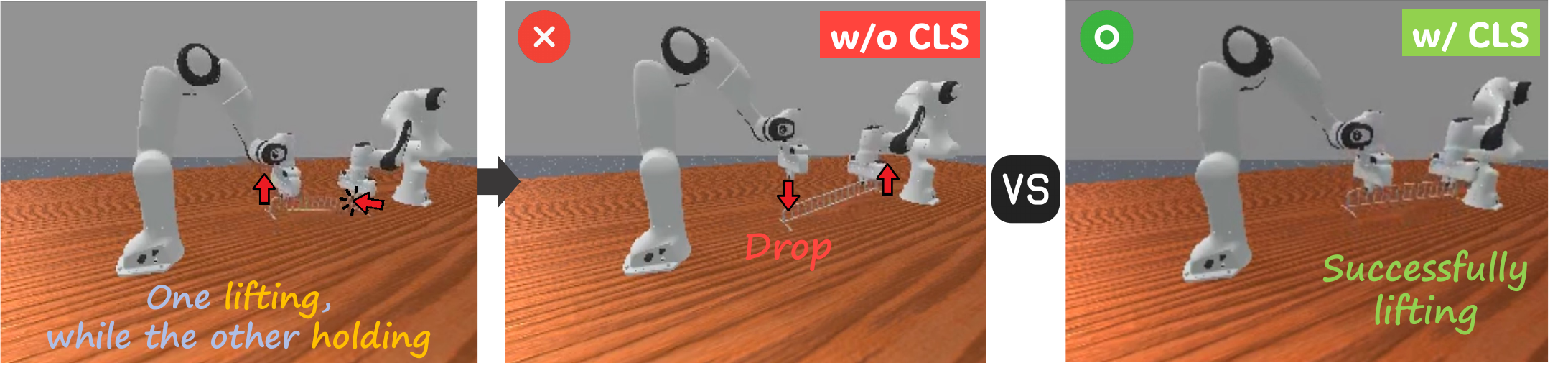}
        \label{fig:fail_1}
    } \vspace{-5pt} \\
    \subfloat[Failure \circled{2}\,: Inter-agent collision from violated sequential dependencies.]{
        \includegraphics[width=\linewidth]{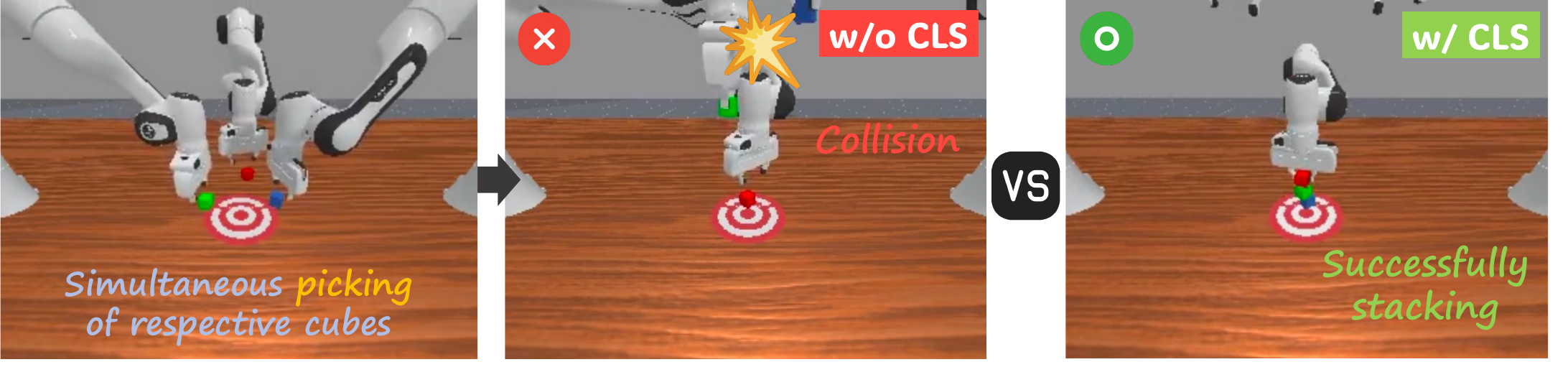}
        \label{fig:fail_2}
    } \vspace{-5pt} \\
    \subfloat[Failure \circled{3}\,: Backward tilt from imbalanced asymmetric lifting.]{
        \includegraphics[width=\linewidth]{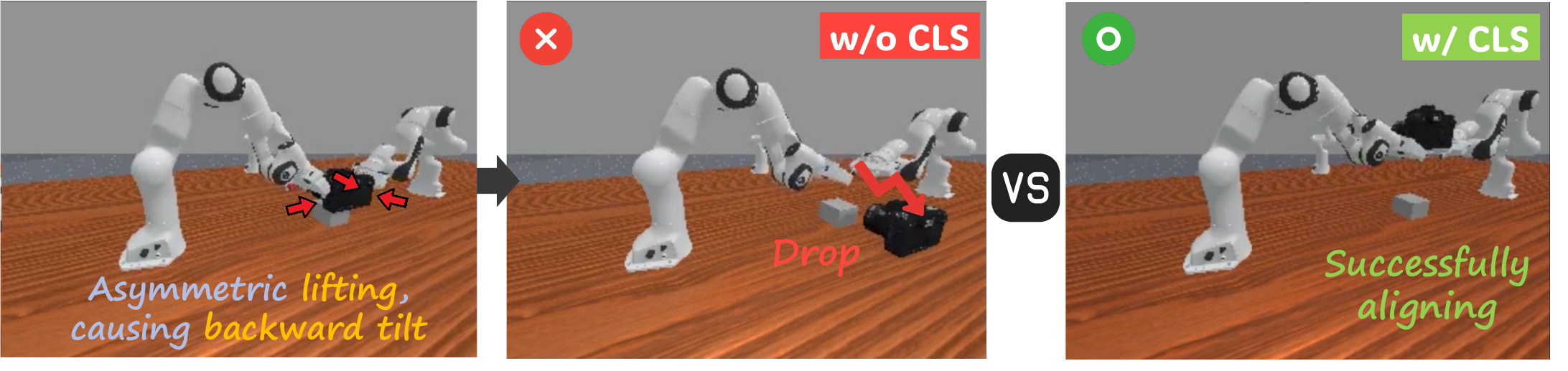}
        \label{fig:fail_3}
    }
    \caption{\hspace{-5pt}Different coordination failures in multi-agent manipulation tasks on RoboFactory~\cite{qin2025robofactory}. Each failure case illustrates a distinct coordination breakdown arising from the absence of the collaborative latent $\mathbf{z}^i_t$, while CLS-DP conditioned on $\mathbf{z}^i_t$ successfully resolves each scenario.}
    \label{fig:failure}
    \vspace{-5pt}
\end{figure}

\begin{table}[t!]
\centering
\caption{Experimental hyperparameter settings for all methods.}
\vspace{-3pt}
\label{tab:setup}
\begin{tabular}{lc @{\hspace{1em}}|@{\hspace{1em}} lc}
\toprule
\textbf{Learning Config.} & \textbf{Value} & \textbf{Execution Config.} & \textbf{Value} \\
\midrule
Learning rate & $10^{-4}$ & Prediction horizon ($H$) & 8 \\
Batch size & 32 & History length ($L$) & 3 \\
Denoising steps ($K$) & 100 & Execution steps & 6 \\
\midrule
\multicolumn{4}{c}{\textbf{Contextualizer Config.}} \\
\midrule
Vision\,/\,Text encoder & ViT-Base & Latent dimension & 256 \\
Feature dimension & 768 & KL weight ($\beta$) & $10^{-1}$ \\
Batch size & 512 & $\beta$ warm-up (linear) & 40\% \\
\bottomrule
\end{tabular}
\vspace{-15pt}
\end{table}

\section{Experiment}
\label{sec:5}

\subsection{Experiment Setup}
Collecting demonstrations for multi-agent manipulation in real-world environments is inherently difficult, as agents must coordinate, synchronize, and physically interact; coordination failures can lead to physical damage to both agents and target objects, as illustrated in Fig.~\ref{fig:failure}.
To avoid these risks, we conduct experiments in simulation on the RoboFactory benchmark~\cite{qin2025robofactory}, selecting six multi-arm tasks spanning two to four agents that cover tight synchronization, role-asymmetric coordination, and strict sequential dependency.

Comparisons are made against diffusion-policy-based and autoregressive BC baselines under both centralized and decentralized execution.
All methods are trained for 100 epochs employing the Adam optimizer with a learning rate warm-up followed by cosine decay.
The policy observation encoder is standardized across all methods, with ResNet-18 and a lightweight multi-layer perceptron (MLP), ensuring a fair comparison across input modalities~\cite{wang2025reinventing}.
Table~\ref{tab:setup} summarizes identical hyperparameter settings across all methods, including the contextualizer configuration.

\BfPara{Centralized Baselines} A unified policy takes all agents' local observations as input and jointly predicts their actions.
\begin{itemize}
    \item \textbf{DP}~\cite{chi2024diffusion} is a vanilla diffusion policy.
    \item \textbf{LargeDP}~\cite{chi2024diffusion} is a parameter-scaled variant of DP to assess the effect of model size.
    \item \textbf{2D Dense Policy}~\cite{su2025dense} adopts bidirectional autoregressive action learning from 2D observations.
    \item \textbf{DP3}~\cite{ze20243d} and \textbf{3D Dense Policy}~\cite{su2025dense} extend DP and 2D Dense Policy to handle 3D point cloud observations.
    \item \textbf{Global GauDP}~\cite{wang2025reinventing} reconstructs a global 3D Gaussian field via 3D Gaussian splatting~\cite{kerbl2023gaussian} from all agents' local observations to form a global scene context.
\end{itemize}

\BfPara{Decentralized Baselines} Each agent executes an agent-specific policy independently under decentralized execution.
\begin{itemize}
    \item \textbf{Local GauDP}~\cite{wang2025reinventing} adopts separate per-agent diffusion policies in contrast to Global GauDP's unified policy, serving as its per-agent baseline.
    \item \textbf{Ours w/o CLS} is a decentralized diffusion policy without the collaborative latent $\mathbf{z}^i_t$ from our contextualizer.
\end{itemize}

\begin{table*}[t!]
  \centering
  \caption{Average success rates across benchmark tasks over 100 unseen episodes. The results are ranked as \colorbox{bestbg}{1st}, \colorbox{secondbg}{2nd}, \colorbox{thirdbg}{3rd}.}
  \vspace{-5pt}
  \label{tab:main_results}
  
  \setlength{\tabcolsep}{4pt} 
  \renewcommand{\arraystretch}{1.05} 
  
  \begin{tabular}{l c  c  c  c  c  c  c} 
    \toprule[1pt]
    
    \multicolumn{1}{c}{\multirow{3}{*}{\textit{Methods}}} & 
    \multicolumn{3}{c}{2 Agents} & 
    \multicolumn{2}{c}{3 Agents} & 
    \multicolumn{1}{c}{4 Agents} &
    \multicolumn{1}{c}{\multirow{3}{*}{Total}} \\ 
    
    \cmidrule(lr){2-4} \cmidrule(lr){5-6} \cmidrule(lr){7-7}
    
    & 
    \multicolumn{1}{c}{Lift Barrier} & 
    \multicolumn{1}{c}{Place Food} & 
    \multicolumn{1}{c}{\begin{tabular}{@{}c@{}} Stack Cube \end{tabular}} & 
    \multicolumn{1}{c}{\begin{tabular}{@{}c@{}} Camera Align. \end{tabular}} & 
    \multicolumn{1}{c}{\begin{tabular}{@{}c@{}} Stack Cube \end{tabular}} & 
    \multicolumn{1}{c}{Take Photo} & \\ \midrule[.5pt]
    \multicolumn{1}{l}{\textit{\textbf{Centralized Policies}}} \\[0.05cm]
    
    DP~\cite{chi2024diffusion} & 
    \resbar[white]{9}{color0} & \resbar[white]{12}{color1} & \resbar[thirdbg]{6}{color2} & \resbar[white]{3}{color3} & \resbar[white]{0}{color4} & \resbar[white]{0}{color5} &
    \resbar[white]{5}{black!20} \\ 
    LargeDP~\cite{chi2024diffusion} & 
    \resbar[thirdbg]{60}{color0} & \resbar[white]{12}{color1} & \resbar[white]{4}{color2} & \resbar[secondbg]{29}{color3} & \resbar[white]{0}{color4} & \resbar[white]{0}{color5} &
    \resbar[thirdbg]{18}{black!20} \\ 
    2D Dense Policy~\cite{su2025dense} & 
    \resbar[white]{3}{color0} & \resbar[white]{2}{color1} & \resbar[white]{0}{color2} & \resbar[white]{0}{color3} & \resbar[white]{0}{color4} & \resbar[secondbg]{9}{color5} &
    \resbar[white]{2}{black!20} \\ 
    DP3 (XYZ)~\cite{ze20243d} & 
    \resbar[white]{30}{color0} & \resbar[thirdbg]{21}{color1} & \resbar[white]{1}{color2} & \resbar[white]{3}{color3} & \resbar[white]{0}{color4} & \resbar[secondbg]{9}{color5} &
    \resbar[white]{11}{black!20} \\ 
    
    DP3 (XYZ+RGB)~\cite{ze20243d} & 
    \resbar[white]{31}{color0} & \resbar[secondbg]{25}{color1} & \resbar[white]{1}{color2} & \resbar[white]{18}{color3} & \resbar[white]{0}{color4} & \resbar[bestbg]{11}{color5} &
    \resbar[white]{14}{black!20} \\ 
    
    3D Dense Policy~\cite{su2025dense} & 
    \resbar[white]{28}{color0} & \resbar[white]{18}{color1} & \resbar[white]{0}{color2} & \resbar[white]{0}{color3} & \resbar[white]{0}{color4} & \resbar[white]{7}{color5} &
    \resbar[white]{9}{black!20} \\
    Global GauDP~\cite{wang2025reinventing} & 
    \resbar[bestbg]{72}{color0} & \resbar[white]{15}{color1} & \resbar[white]{2}{color2} & \resbar[thirdbg]{26}{color3} & \resbar[white]{0}{color4} & \resbar[white]{3}{color5} &
    \resbar[secondbg]{20}{black!20} \\ \midrule[.5pt]
    \multicolumn{1}{l}{\textit{\textbf{Decentralized Policies}}} \\[0.05cm]
    Local GauDP~\cite{wang2025reinventing} & 
    \resbar[white]{3}{color0} & \resbar[white]{12}{color1} & \resbar[white]{0}{color2} & \resbar[white]{15}{color3} & \resbar[white]{0}{color4} & \resbar[white]{2}{color5} &
    \resbar[white]{5}{black!20} \\
    
    \textbf{CLS-DP (Ours)} & 
    \resbar[secondbg]{61}{color0} & 
    \resbar[bestbg]{43}{color1} & 
    \resbar[bestbg]{39}{color2} & 
    \resbar[bestbg]{55}{color3} & 
    \resbar[bestbg]{20}{color4} & 
    \resbar[thirdbg]{8}{color5} &
    \resbar[bestbg]{38}{black!40} \\ 
    
    Ours w/o CLS & 
    \resbar[white]{14}{color0} & \resbar[white]{5}{color1} & \resbar[secondbg]{14}{color2} & \resbar[white]{7}{color3} & \resbar[secondbg]{8}{color4} & \resbar[white]{3}{color5} &
    \resbar[white]{9}{black!20} \\
    
    \bottomrule[1pt]
  \end{tabular}
  \vspace{-15pt}
\end{table*}

\subsection{Quantitative Results}

\subsubsection{Task Success Rate}
\label{sec:5-B-1}
Table~\ref{tab:main_results} reports the average success rate across all methods, evaluated over 100 episodes with unseen seeds that introduce varied target object placements.

\BfPara{Centralized Baselines}
Global GauDP achieves the highest average success rate across all tasks (20\%) among centralized baselines, demonstrating the benefit of reconstructing a shared 3D Gaussian field via Gaussian splatting from all agents' local observations.
However, its performance decreases sharply as tasks involve more agents (72\% on \textit{Lift Barrier}, 26\% on \textit{Camera Alignment}, and 3\% on \textit{Take Photo}). It also remains weak on sequential stacking tasks (2\% on \textit{Two Robots Stack Cube} and 0\% on \textit{Three Robots Stack Cube}).

For methods that directly take 3D point cloud observations as input, DP3 (XYZ+RGB) performs best on \textit{Take Photo} (11\%), where 3D geometric information appears to help; yet their overall success rates remain low (9--14\%). In addition, both DP3 and 3D Dense Policy are weak on sequential stacking tasks (0--1\% on \textit{Two} and \textit{Three Robots Stack Cube}), which require precise subgoal sequencing and inter-agent synchronization. They also show limited performance on \textit{Lift Barrier} (28--31\%) and \textit{Camera Alignment} (0--18\%).
This suggests that irrelevant global features can dilute locally critical cues and hinder local responsiveness.

Among 2D-modality methods, DP (5\%) and 2D Dense Policy (2\%) perform poorly overall. LargeDP ranks second with 18\% on average by increasing model size, yet records 0\% on tasks with more agents, such as \textit{Three Robots Stack Cube} and \textit{Take Photo}.
It also achieves low success rates on two-agent tasks, including \textit{Place Food} (12\%) and \textit{Two Robots Stack Cube} (4\%).
This reveals that model scale alone is insufficient for effective multi-agent manipulation.

\BfPara{Decentralized Baselines}
Local GauDP (5\%) controls each agent independently rather than leveraging a unified global policy for modeling inter-agent dependencies, and exhibits severe performance degradation.
Ours w/o CLS (9\%), a decentralized diffusion policy without the collaborative latent $\mathbf{z}^i_t$, similarly struggles on tasks requiring tight inter-agent coordination.
The substantial gap between CLS-DP and Ours w/o CLS offers a quantitative measure of the contribution of the collaborative latent to implicit coordination.

\BfPara{CLS-DP}
Our CLS-DP achieves a total average success rate of 38\%, outperforming all centralized and decentralized baselines despite each agent relying only on its own local observation and the shared task instruction at deployment.
It achieves substantial gains across nearly all tasks, including \textit{Place Food} (43\%), \textit{Two Robots Stack Cube} (39\%), and \textit{Camera Alignment} (55\%), and is the only method to reach non-trivial performance on \textit{Three Robots Stack Cube} (20\%).

However, CLS-DP ranks third in \textit{Take Photo} (8\%) because methods with explicit spatial inputs (e.g., 3D point clouds or dense 2D representations) can leverage more direct geometric information for the final shutter press. In contrast, CLS-DP implicitly infers coordination-relevant context at the latent level from only 2D local observation, incurring a lower computational cost.
This suggests that the modest gap (8\% vs. 11\%) from the best baseline on this task, DP3 (XYZ+RGB), stems from fine-grained spatial precision rather than missing coordination cues.
It also outperforms both Global GauDP (3\%) and Local GauDP (2\%), indicating that latent-space distillation mitigates coordination failures more effectively than global scene reconstruction, when fine-grained spatial precision remains a bottleneck.

Overall, CLS-DP provides coordination cues while preserving per-agent local control. This yields nearly twice the success rate of the best centralized baseline (38\% vs. 20\%), with improved scalability to more agents.

\subsubsection{Parameter Efficiency and Scalability}
As shown in Table~\ref{tab:param_efficiency_v3}, we evaluate parameter efficiency as the success rate per parameter count across 2--4 agent configurations.
For GauDP, the best-performing centralized baseline, we distinguish two variants: \textit{i)} the full deployment model, which executes the Gaussian reconstruction module at every inference timestep to build a shared 3D scene representation from multi-view RGB local observations; \textit{ii)} the policy-only variant (GauDP-G) without this reconstruction module.
For CLS-DP, we additionally include a conservative variant (CLS-DP+K) that incorporates the multi-agent kinematics encoder--decoder pair employed only during centralized training.

Table~\ref{tab:param_efficiency_v3} shows that CLS-DP achieves the highest efficiency across all settings by a substantial margin, whereas LargeDP increases model capacity but yields lower efficiency than even DP in the 2-agent setting. This confirms that scaling model size alone does not translate to better coordination.
The efficiency gap between GauDP and GauDP-G highlights the perceptual overhead introduced by the reconstruction module, which is required at deployment in GauDP.
In contrast, CLS-DP discards training-only modules at deployment, and the negligible difference between CLS-DP+K and CLS-DP indicates that these modules introduce only a small parameter overhead.
Even under this conservative comparison, CLS-DP+K maintains overwhelming efficiency advantages over GauDP across all agent configurations. These results imply that our performance gains stem from the coordination mechanism rather than perceptual overhead.
\vspace{-3pt}

\begin{table}[h!]
\vspace{-3pt}
\centering
\caption{\textbf{Efficiency Comparisons} ($\uparrow$):\vspace{-2pt}
\\Average Success Rate ($\uparrow$) [\%] per Parameters ($\downarrow$) [M].}
\vspace{-10pt}
\label{tab:param_efficiency_v3}

\renewcommand{\tabularxcolumn}[1]{m{#1}}

\begin{tabularx}{\columnwidth}{l | X | X | X}
\toprule[1pt]
\multicolumn{1}{c|}{\textbf{Method}} &
\multicolumn{1}{c|}{\textbf{2 Agents}} & 
\multicolumn{1}{c|}{\textbf{3 Agents}} & 
\multicolumn{1}{c}{\textbf{4 Agents}} \\
\midrule[.5pt]

DP~\cite{chi2024diffusion}
& \cellcolor{thirdbg}$0.0693$ \hfill \tikz[baseline=-0.5ex]{
    \draw[gray,line width=.3pt] (0,0) -- (0.4,0); 
    \draw[black,line width=1.5pt] (0,0) -- (0.135,0); }
& $0.0092$ \hfill \tikz[baseline=-0.5ex]{ 
    \draw[gray,line width=.3pt] (0,0) -- (0.4,0); 
    \draw[black,line width=1.5pt] (0,0) -- (0.032,0); }
& $0.0000$ \hfill \tikz[baseline=-0.5ex]{ 
    \draw[gray,line width=.3pt] (0,0) -- (0.4,0); 
    \draw[black,line width=1.5pt] (0,0) -- (0.0,0); } \\

LargeDP~\cite{chi2024diffusion}
& $0.0351$ \hfill \tikz[baseline=-0.5ex]{ 
    \draw[gray,line width=.3pt] (0,0) -- (0.4,0); 
    \draw[black,line width=1.5pt] (0,0) -- (0.068,0); }
& $0.0152$ \hfill \tikz[baseline=-0.5ex]{ 
    \draw[gray,line width=.3pt] (0,0) -- (0.4,0); 
    \draw[black,line width=1.5pt] (0,0) -- (0.053,0); }
& $0.0000$ \hfill \tikz[baseline=-0.5ex]{ 
    \draw[gray,line width=.3pt] (0,0) -- (0.4,0); 
    \draw[black,line width=1.5pt] (0,0) -- (0.0,0); } \\

GauDP-G~\cite{wang2025reinventing}
& $0.0395$ \hfill \tikz[baseline=-0.5ex]{ 
    \draw[gray,line width=.3pt] (0,0) -- (0.4,0); 
    \draw[black,line width=1.5pt] (0,0) -- (0.077,0); }
& \cellcolor{thirdbg}$0.0166$ \hfill \tikz[baseline=-0.5ex]{ 
    \draw[gray,line width=.3pt] (0,0) -- (0.4,0); 
    \draw[black,line width=1.5pt] (0,0) -- (0.058,0); }
& \cellcolor{thirdbg}$0.0037$ \hfill \tikz[baseline=-0.5ex]{ 
    \draw[gray,line width=.3pt] (0,0) -- (0.4,0); 
    \draw[black,line width=1.5pt] (0,0) -- (0.078,0); } \\

GauDP~\cite{wang2025reinventing}
& $0.0068$ \hfill \tikz[baseline=-0.5ex]{ 
    \draw[gray,line width=.3pt] (0,0) -- (0.4,0); 
    \draw[black,line width=1.5pt] (0,0) -- (0.013,0); }
& $0.0035$ \hfill \tikz[baseline=-0.5ex]{ 
    \draw[gray,line width=.3pt] (0,0) -- (0.4,0); 
    \draw[black,line width=1.5pt] (0,0) -- (0.012,0); }
& $0.0009$ \hfill \tikz[baseline=-0.5ex]{ 
    \draw[gray,line width=.3pt] (0,0) -- (0.4,0); 
    \draw[black,line width=1.5pt] (0,0) -- (0.019,0); } \\

\midrule[.5pt]

CLS-DP+K (ours)
& \cellcolor{secondbg}$0.2013$ \hfill \tikz[baseline=-0.5ex]{ 
    \draw[gray,line width=.3pt] (0,0) -- (0.4,0); 
    \draw[black,line width=1.5pt] (0,0) -- (0.391,0); }
& \cellcolor{secondbg}$0.1124$ \hfill \tikz[baseline=-0.5ex]{ 
    \draw[gray,line width=.3pt] (0,0) -- (0.4,0); 
    \draw[black,line width=1.5pt] (0,0) -- (0.392,0);  }
& \cellcolor{secondbg}$0.0186$ \hfill \tikz[baseline=-0.5ex]{ 
    \draw[gray,line width=.3pt] (0,0) -- (0.4,0); 
    \draw[black,line width=1.5pt] (0,0) -- (0.392,0); } \\

\textbf{CLS-DP (ours)}
& \cellcolor{bestbg}$\mathbf{0.2056}$ \hfill \tikz[baseline=-0.5ex]{ 
    \draw[gray,line width=.3pt] (0,0) -- (0.4,0); 
    \draw[black,line width=1.5pt] (0,0) -- (0.4,0); }
& \cellcolor{bestbg}$\mathbf{0.1148}$ \hfill \tikz[baseline=-0.5ex]{ 
    \draw[gray,line width=.3pt] (0,0) -- (0.4,0); 
    \draw[black,line width=1.5pt] (0,0) -- (0.4,0); }
& \cellcolor{bestbg}$\mathbf{0.0190}$ \hfill \tikz[baseline=-0.5ex]{ 
    \draw[gray,line width=.3pt] (0,0) -- (0.4,0); 
    \draw[black,line width=1.5pt] (0,0) -- (0.4,0); } \\

\bottomrule[1pt]
\end{tabularx}
\vspace{-5pt}
\end{table}

\begin{table}[h!]
\caption{Examples of text instructions per task.}
\vspace{-8pt}
\label{tab:instructions}
\centering
\scriptsize
\setlength{\tabcolsep}{6pt}
\renewcommand{\arraystretch}{1.3}
\begin{tabular}{|p{0.9\columnwidth}|}
\hline
\rowcolor{gray!20} \textbf{Lift Barrier} \\
{\tiny$\bullet$} Lift the metal barrier and keep it straight. \\
{\tiny$\bullet$} Grasp the metal barrier firmly and raise it to the target height. \\
\hline
\rowcolor{gray!20} \textbf{Place Food} \\
{\tiny$\bullet$} Lift the pot lid and place a small piece of food inside. \\
{\tiny$\bullet$} Open the lid and move the food item to the center of the pot. \\
\hline
\rowcolor{gray!20} \textbf{Two Robots Stack Cube} \\
{\tiny$\bullet$} Move the blue cube to the target and stack the red cube on top. \\
{\tiny$\bullet$} Position the blue cube and stack the red cube. \\
\hline
\rowcolor{gray!20} \textbf{Camera Alignment} \\
{\tiny$\bullet$} Place the object at its target position and raise the camera to match. \\
{\tiny$\bullet$} Hold the object at its target position and align the camera precisely. \\
\hline
\rowcolor{gray!20} \textbf{Three Robots Stack Cube} \\
{\tiny$\bullet$} Place the blue cube, then stack the red and green cubes on top. \\
{\tiny$\bullet$} Position the blue cube and carefully place the red and green cubes. \\
\hline
\rowcolor{gray!20} \textbf{Take Photo} \\
{\tiny$\bullet$} Move the object to the target, align the camera, then press the shutter. \\
{\tiny$\bullet$} Place the object, align the camera, and press the shutter afterward. \\
\hline
\end{tabular}
\vspace{-5pt}
\end{table}

\addtocounter{figure}{1}
\begin{figure*}[t!]
    \centering
    \includegraphics[width=\linewidth]{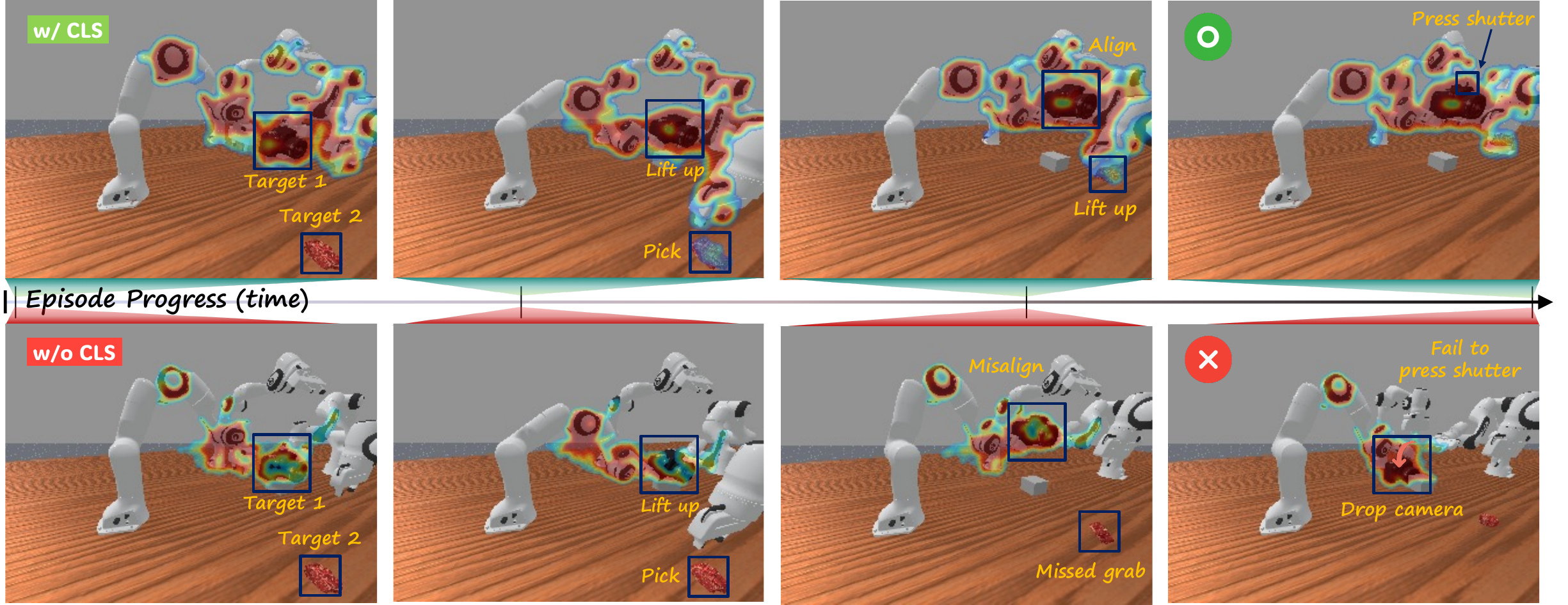}
    \caption{\hspace{-5pt}\textbf{Attribution analysis via Integrated Gradients~\cite{sundararajan2017axiomatic} for the \textit{Take Photo} task.} Attribution maps highlight regions of the local image observation that most influence the predicted action sequence over time. CLS-DP (top) consistently shifts attribution not only to its own joints and gripper but also to those of other agents as execution progresses, successfully completing the task. In contrast, the baseline diffusion policy without $\mathbf{z}^i_t$ (bottom) produces sparse, egocentric maps concentrated around its own joints and gripper, failing to track teammates, which leads to a misaligned and dropped camera (target 1) as well as a missed grab. These results demonstrate that $\mathbf{z}^i_t$ successfully captures both inter-agent coordination dynamics and the agent's own task progression.}
    \label{fig:attention_analysis}
    \vspace{-10pt}
\end{figure*}

\addtocounter{figure}{-2}
\begin{figure}[h!]
    \centering
    \includegraphics[width=\columnwidth]{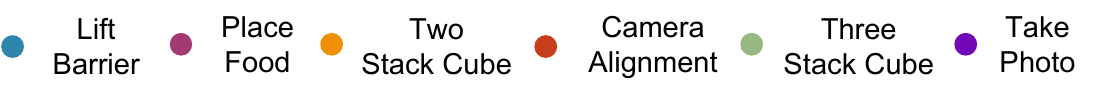}\\
    \includegraphics[width=\columnwidth]{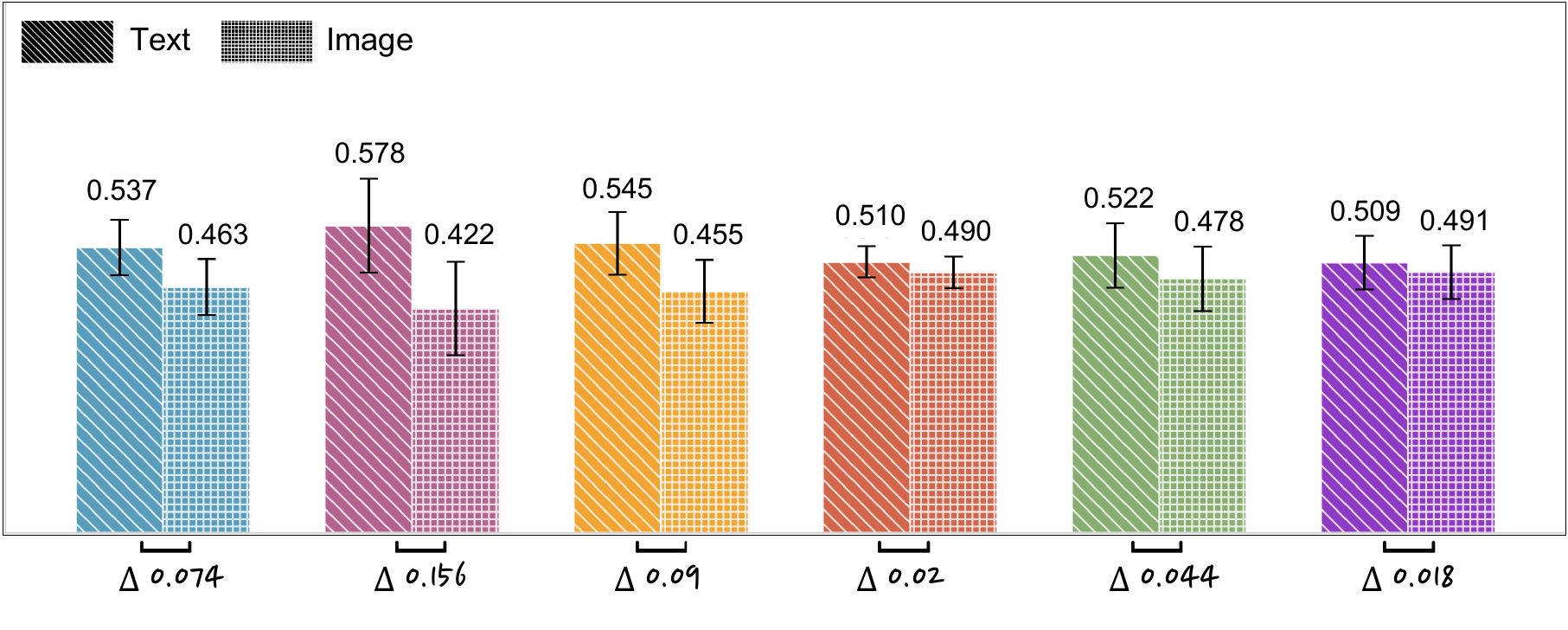}
    \hspace{1mm}
    \vspace{-20pt}
    \caption{\hspace{-5pt}Task-wise analysis of cross-attention weights in the contextualizer. $\Delta$ denotes the text–image attention gap for each task.}
    \vspace{-20pt}
    \label{fig:weight_analysis}
\end{figure}

\subsubsection{Cross-Modal Grounding Analysis}
Fig.~\ref{fig:weight_analysis} reports the cross-attention weights between text instructions and image observations in the contextualizer across all tasks.
For each task, we utilize a large language model (LLM) to generate task-level instructions from the task and object descriptions provided by RoboFactory~\cite{qin2025robofactory}.
Following RoboTwin 2.0~\cite{chen2025robotwin2}, we diversify instructions by sampling from LLM-generated alternative phrasings, producing 100 instructions for training and 100 held-out instructions for evaluation.
At each episode, a sampled instruction is shared across all agents. Table~\ref{tab:instructions} shows examples of these instructions.

Both modalities contribute across all tasks, revealing that the contextualizer grounds the collaborative latent from both image observations and text instructions.
Throughout all tasks, each contextualizer relies slightly but consistently more on textual context than image observations for task-level coordination, though the relative weighting between modalities varies with task demands.

\textit{Lift Barrier} shows a text-heavy pattern ($\Delta$ = 0.074), where the instruction specifies goal constraints (e.g., keeping the barrier straight and reaching a target height) while image observations provide corrective feedback during lifting.
\textit{Place Food} shows the largest gap ($\Delta$ = 0.156), consistent with role-asymmetric instructions that explicitly specify the key subgoals and their order (e.g., opening the lid before placing food).
The cube-stacking tasks also rely on sequential decomposition specified by text (e.g., placing the base cube before stacking others), though the gap narrows from \textit{Two Robots Stack Cube} ($\Delta$ = 0.09) to \textit{Three Robots Stack Cube} ($\Delta$ = 0.044).
The smaller gap in \textit{Three Robots Stack Cube} is consistent with increased reliance on image feedback as the number of stacking steps grows, requiring more frequent verification and corrective adjustments during execution.
\textit{Camera Alignment} and \textit{Take Photo} show the smallest gaps ($\Delta$ = 0.02 and 0.018), as camera-to-target matching relies more on visual feedback.

Overall, these trends confirm that the contextualizer adaptively modulates cross-modal attention according to the coordination demands of each task.
\vspace{-5pt}

\subsection{Qualitative Analysis}
Fig.~\ref{fig:attention_analysis} visualizes Integrated Gradients~\cite{sundararajan2017axiomatic} attribution maps for the \textit{Take Photo} task, which highlight the regions of the local image observation that most influence the predicted action sequence throughout the episode.

At the initial timestep, the agent trained with CLS-DP assigns high attribution not only to its own joints and gripper but also to those of other agents engaged in the same task.
As execution progresses, attribution remains concentrated on these regions, reflecting sensitivity to both its own and its teammates' kinematics.
The temporally coherent attribution patterns indicate that the collaborative latent captures the progression of teammates' dynamics without explicit communication throughout execution. These patterns confirm that coordination cues are well encoded in \textit{Take Photo}.

In contrast, the baseline without $\mathbf{z}^i_t$ produces sparse and egocentric attribution maps; this leads to a coordination failure in which the camera (target 1) is misaligned and ultimately dropped.
The baseline also fails to grab the object (target 2)---a task requiring no inter-agent coordination---whereas CLS-DP succeeds. This suggests that the collaborative latent distills the agent's own task progression as well as the dynamics of other agents, which can benefit individual task execution beyond inter-agent coordination alone.

%% file: Paper/6_conclusion.tex
\section{Conclusion}
\label{sec:6}
We presented CLS-DP, a decentralized multi-agent framework in which each agent infers collaborative dynamics from per-agent local RGB observation and a shared task instruction.
CLS-DP learns a collaborative latent space by distilling privileged multi-agent joint trajectories available only during training.
By conditioning the diffusion denoising process on this latent, each agent can infer teammates' dynamics and coordinate under partial observability without shared global observations, explicit global state information, or inter-agent communication.
Furthermore, per-agent cost remains independent of team size, thereby improving scalability to larger teams.
We validate CLS-DP on six RoboFactory tasks, where it outperforms all centralized and decentralized baselines with higher parameter efficiency.
Even though latent-level coordination inference introduces a fine-grained spatial precision bottleneck, CLS-DP achieves performance competitive with methods that directly employ geometric global scene information, at lower computational cost by relying on 2D local observation.

Future directions include extending CLS-DP to real-world deployment by integrating safety constraints such as collision avoidance, and exploring its applicability to more diverse multi-agent environments beyond manipulation tasks.

%% file: ref.bib
@inproceedings{he2025latent,
  title={Latent Theory of Mind: A Decentralized Diffusion Architecture for Cooperative Manipulation},
  author={He, Chengyang and others},
  booktitle={CoRL},
  publisher = {PMLR},
  year={2025},
}

@inproceedings{su2025dense,
  title={Dense Policy: Bidirectional Autoregressive Learning of Actions},
  author={Su, Yue and others},
  booktitle={ICCV},
  publisher = {IEEE},
  year={2025},
  pages={14486-14495},
}

@inproceedings{ze20243d,
  title={{3D} Diffusion Policy: Generalizable Visuomotor Policy Learning via Simple {3D} Representations},
  author={Ze, Yanjie and others},
  booktitle={RSS},
  volume={20},
  year={2024},
}

@inproceedings{wang2025reinventing,
  title={{GauDP}: Reinventing Multi-Agent Collaboration through Gaussian-Image Synergy in Diffusion Policies},
  author={Wang, Ziye and others},
  booktitle={NeurIPS},
  pages = {5620--5639},
  year={2025},
}

@inproceedings{chi2024diffusion,
  title={Diffusion Policy: Visuomotor Policy Learning via Action Diffusion},
  author={Chi, Cheng and others},
  booktitle={RSS},
  volume={19},
  year={2023},
}

@book{worldrobotics2025,
  author    = {M{\"u}ller, Christopher},
  title     = {World Robotics 2025: Industrial Robots},
  year      = {2025},
  publisher = {VDMA Services GmbH},
  institution = {IFR Statistical Department},
}

@inproceedings{kim2024srt,
  author    = {Kim, J. W. and others},
  title     = {Surgical Robot Transformer ({SRT}): Imitation Learning for Surgical Tasks},
  booktitle = {CoRL},
  publisher = {PMLR},
  year      = {2024},
  pages     = {130-144}
}

@inproceedings{tang2024automate,
  author    = {Tang, Bingjie and others},
  title     = {{AutoMate}: Specialist and Generalist Assembly Policies over Diverse Geometries},
  booktitle = {RSS},
  volume   = {20},
  year      = {2024},
}

@inproceedings{chi2024umi,
  author    = {Chi, Cheng and others},
  title     = {Universal Manipulation Interface: In-The-Wild Robot Teaching Without In-The-Wild Robots},
  booktitle = {RSS},
  volume   = {20},
  year      = {2024},
}

@inproceedings{nasiriany2024robocasa,
  author    = {Nasiriany, Soroush and others},
  title     = {{RoboCasa}: Large-Scale Simulation of Household Tasks for Generalist Robots},
  booktitle = {RSS},
  volume   = {20},
  year      = {2024},
}

@inproceedings{jiang2023motiondiffuser,
  title={{MotionDiffuser}: Controllable Multi-Agent Motion Prediction using Diffusion},
  author={Jiang, Chiyu and others},
  booktitle = {CVPR},
  publisher = {IEEE},
  year = {2023},
  pages={9644-9653}
}

@inproceedings{dong2025mimicd,
    author = {Dong, Dayi and others},
    title = {{MIMIC-D}: Multi-modal Imitation for Multi-Agent Coordination with Decentralized Diffusion Policies},
    booktitle = {ICRA},
    year = {2026},
    publisher = {IEEE}
}

@inproceedings{ho2020denoising,
  title={Denoising Diffusion Probabilistic Models},
  author={Ho, Jonathan and others},
  booktitle={NeurIPS},
  pages={6840--6851},
  year={2020},
}

@inproceedings{niedoba2023diffusion,
  title={A Diffusion-Model of Joint Interactive Navigation},
  author={Niedoba, Matthew and others},
  booktitle={NeurIPS},
  pages={55995--56011},
  year={2023}
}

@inproceedings{zhu2024madiff,
  title={{MADiff}: Offline Multi-agent Learning with Diffusion Models},
  author={Zhu, Zhengbang and others},
  booktitle={NeurIPS},
  pages = {4177--4206},
  year={2024}
}

@inproceedings{qin2025robofactory,
  title={{RoboFactory}: Exploring Embodied Agent Collaboration with Compositional Constraints},
  author={Qin, Yiran and others},
  booktitle={ICCV},
  publisher = {IEEE},
  pages={10075-10085},
  year={2025}
}

@inproceedings{sundararajan2017axiomatic,
  title={Axiomatic Attribution for Deep Networks},
  author={Sundararajan, Mukund and others},
  booktitle={ICML},
  publisher = {PMLR},
  pages={3319--3328},
  year={2017}
}

@inproceedings{dalal2023imitating,
  title={Imitating Task and Motion Planning with Visuomotor Transformers},
  author={Dalal, Murtaza and others},
  booktitle={CoRL},
  publisher = {PMLR},
  pages={2565--2593},
  year={2023}
}

@inproceedings{ma2024contrastive,
  title={Contrastive Imitation Learning for Language-guided Multi-Task Robotic Manipulation},
  author={Ma, Teli and others},
  booktitle={CoRL},
  publisher = {PMLR},
  pages={4651--4669},
  year={2024}
}

@inproceedings{foster2024is,
  title={Is Behavior Cloning All You Need? Understanding Horizon in Imitation Learning},
  author={Foster, Dylan J. and others},
  booktitle={NeurIPS},
  pages = {120602--120666},
  year={2024}
}

@inproceedings{yin2024offline,
  title={Offline Imitation Learning Through Graph Search and Retrieval},
  author={Yin, Zhao-Heng and others},
  booktitle={RSS},
  volume={20},
  year={2024}
}

@inproceedings{hansen2024tdmpc2,
  title={{TD-MPC2}: Scalable, Robust World Models for Continuous Control},
  author={Hansen, Nicklas and others},
  booktitle={ICLR},
  year={2024}
}

@inproceedings{zhou2025dinowm,
  title={{DINO-WM}: World Models on Pre-trained Visual Features enable Zero-shot Planning},
  author={Zhou, Gaoyue and others},
  booktitle={ICML},
  publisher = {PMLR},
  pages = {79115--79135},
  year={2025}
}

@inproceedings{cui2024dynamo,
  title={{DynaMo}: In-Domain Dynamics Pretraining for Visuo-Motor Control},
  author={Cui, Zichen Jeff and others},
  booktitle={NeurIPS},
  pages = {33933--33961},
  year={2024}
}

@inproceedings{hansen2025hierarchical,
  title={Hierarchical World Models as Visual Whole-Body Humanoid Controllers},
  author={Hansen, Nicklas and others},
  booktitle={ICLR},
  year={2025}
}

@inproceedings{perez2018film,
  title={{FiLM}: Visual Reasoning with a General Conditioning Layer},
  author={Perez, Ethan and others},
  booktitle={AAAI},
  pages={3942--3951},
  volume={32},
  year={2018}
}

@inproceedings{florence2021implicit,
  title={Implicit Behavioral Cloning},
  author={Florence, Pete and others},
  booktitle={CoRL},
  publisher = {PMLR},
  pages={158--168},
  year={2021}
}

@inproceedings{du2021improved,
  title={Improved Contrastive Divergence Training of Energy-Based Models},
  author={Du, Yilun and others},
  booktitle={ICML},
  pages={2837--2848},
  year={2021},
  publisher={PMLR}
}

@inproceedings{ta2022conditional,
  title={Conditional Energy-Based Models for Implicit Policies: The Gap between Theory and Practice},
  author={Ta, Duy-Nguyen and others},
  booktitle={{IMRSS}: Workshop on Implicit Representations for Robotic Manipulation @ RSS},
  year={2022}
}

@inproceedings{song2019generative,
  title={Generative modeling by estimating gradients of the data distribution},
  author={Song, Yang and others},
  booktitle={NeurIPS},
  year={2019},
  pages={11895--11907}
}

@inproceedings{wang2022diffusion,
  title={Diffusion policies as an expressive policy class for offline reinforcement learning},
  author={Wang, Zhendong and others},
  booktitle={ICLR},
  year={2023}
}

@article{amato2024initial,
  title={An Initial Introduction to Cooperative Multi-Agent Reinforcement Learning},
  author={Amato, Christopher},
  journal={arXiv preprint arXiv:2405.06161},
  year={2024}
}

@book{oliehoek2016concise,
  title={A Concise Introduction to Decentralized {POMDPs}},
  author={Oliehoek, Frans A. and others},
  year={2016},
  publisher={Springer},
}

@article{xue2025leverb,
  title={{LeVERB}: Humanoid Whole-Body Control with Latent Vision-Language Instruction},
  author={Xue, Haoru and others},
  journal={arXiv preprint arXiv:2506.13751},
  year={2025}
}

@inproceedings{zhai2023sigmoid,
  title={Sigmoid Loss for Language Image Pre-Training},
  author={Zhai, Xiaohua and others},
  booktitle={ICCV},
  publisher = {IEEE},
  pages={11941--11952},
  year={2023}
}

@inproceedings{vaswani2017attention,
  title={Attention is All you Need},
  author={Vaswani, Ashish and others},
  booktitle={NeurIPS},
  pages={5998--6008},
  year={2017}
}

@article{ling2020character,
  title={Character Controllers Using Motion {VAE}s},
  author={Ling, Hung Yu and others},
  journal={ACM Trans. Graph.},
  volume={39},
  number={4},
  pages={1--12},
  year={2020}
}

@inproceedings{parisi2022unsurprising,
  title={The Unsurprising Effectiveness of Pre-Trained Vision Models for Control},
  author={Parisi, Simone and others},
  booktitle={ICML},
  publisher = {PMLR},
  pages={17359--17371},
  year={2022}
}

@inproceedings{gupta2024pretrained,
  title={Pre-trained Text-to-Image Diffusion Models Are Versatile Representation Learners for Control},
  author={Gupta, Gunshi and others},
  booktitle={NeurIPS},
  pages = {74182--74210},
  year={2024}
}

@inproceedings{alemi2018fixing,
  title={Fixing a Broken {ELBO}},
  author={Alemi, Alexander A. and others},
  booktitle={ICML},
  pages={159--168},
  year={2018},
  publisher={PMLR},
}

@article{kerbl2023gaussian,
  title   = {{3D} Gaussian Splatting for Real-Time Radiance Field Rendering},
  author  = {Kerbl, Bernhard and others},
  journal = {ACM Trans. Graph.},
  volume  = {42},
  number  = {4},
  pages   = {1--14},
  year    = {2023},
}

@article{chen2025robotwin2,
  title   = {{RoboTwin} 2.0: A Scalable Data Generator and Benchmark with Strong Domain Randomization for Robust Bimanual Robotic Manipulation},
  author  = {Tianxing Chen and others},
  year    = {2025},
  journal = {arXiv preprint arXiv:2506.18088}
}
